\documentclass{article}

\usepackage{arxiv}

\usepackage[utf8]{inputenc} 
\usepackage[T1]{fontenc}    
\usepackage{hyperref}       
\usepackage{url}            
\usepackage{booktabs}       
\usepackage{amsfonts}       
\usepackage{nicefrac}       
\usepackage{microtype}      
\usepackage{lipsum}
\usepackage{cite}
\usepackage{amsmath,amssymb,amsfonts}
\usepackage{algorithmic}
\usepackage{graphicx}
\usepackage{textcomp}
\usepackage{tikz}
\usepackage{caption}
\usepackage{enumitem}
\usepackage{subfigure}
\usepackage{booktabs}
\usepackage{bm}
\usepackage{hyperref}

\title{Introducing the Hidden Neural Markov Chain framework}

\author{
Elie Azeraf\thanks{Elie Azeraf is also a member of SAMOVAR, Telecom SudParis, Institut Polytechnique de Paris} \\
  Watson Department \\
  IBM GSB France \\
  \texttt{elie.azeraf@ibm.com} \\
   \And
Emmanuel Monfrini \\
SAMOVAR, Telecom SudParis \\
Institut Polytechnique de Paris \\
\And
Emmanuel Vignon \\
Watson Department \\
  IBM GSB France \\
\And
Wojciech Pieczynski \\
SAMOVAR, Telecom SudParis \\
Institut Polytechnique de Paris \\
}

\begin{document}
\maketitle

\begin{abstract}
Nowadays, neural network models achieve state-of-the-art results in many areas as computer vision or speech processing. For sequential data, especially for Natural Language Processing (NLP) tasks, Recurrent Neural Networks (RNNs) and their extensions, the Long Short Term Memory (LSTM) network and the Gated Recurrent Unit (GRU), are among the most used models, having a “term-to-term" sequence processing.
However, if many works create extensions and improvements of the RNN, few have focused on developing other ways for sequential data processing with neural networks in a “term-to-term" way. 
This paper proposes the original Hidden Neural Markov Chain (HNMC) framework, a new family of sequential neural models. They are not based on the RNN but on the Hidden Markov Model (HMM), a probabilistic graphical model. This neural extension is possible thanks to the recent Entropic Forward-Backward algorithm for HMM restoration. We propose three different models: the classic HNMC, the HNMC2, and the HNMC-CN. After describing our models' whole construction, we compare them with classic RNN and Bidirectional RNN (BiRNN) models for some sequence labeling tasks: Chunking, Part-Of-Speech Tagging, and Named Entity Recognition. For every experiment, whatever the architecture or the embedding method used, one of our proposed models has the best results. It shows this new neural sequential framework's potential, which can open the way to new models, and might eventually compete with the prevalent BiLSTM and BiGRU.
\end{abstract}


\section{\uppercase{Introduction}}

During the last years, neural networks models \cite{Goodfellow-et-al-2016, lecun2015deep} show impressive performances in many areas, as computer vision or speech processing.
Among them, Natural Language Processing (NLP) has one of the most significant expansions. 
The Recurrent Neural Network \cite{rumelhart1985learning,jordan1990attractor,jozefowicz2015empirical} (RNN) based models, treating text as sequential data, are among the most often used models for NLP tasks, especially the Long Short Term Memory network (LSTM) \cite{hochreiter1997long} and the Gated Recurrent Unit (GRU) \cite{chung2014empirical}. 
They can cover all textual applications as word embedding \cite{akbik2018coling} or text translation \cite{sutskever2014sequence}. They are the most prevalent sequential models with neural networks, having a term-to-term data processing. 

However, if many works have been done to create extensions of the RNN, very few of them focused on a different way to use neural networks to treat sequential data with term-to-term processing. There are Transformer \cite{vaswani2017attention} based models, as BERT \cite{devlin2018bert} or XLNet \cite{yang2019xlnet}, but they have a different structure as they catch all the observations of the sequence in one time (under padding limitations) and require many more parameters and training power. In this paper, we only focus on neural models with term-to-term processing.

Among the sequential models, one of the most popular is the Hidden Markov Model (HMM) \cite{stratonovich1965conditional,baum1966statistical,rabiner1986introduction}, also called Hidden Markov Chain, which is a probabilistic graphical model \cite{koller2009probabilistic}. In this paper, we propose a new framework of sequential neural models based on HMM, named Hidden Neural Markov Chains (HNMCs), composed of the classic HNMC, the HNMC of order 2 (HNMC2), and the HNMC with complexified noise (HNMC-CN). As RNN, they are neural term-to-term models for sequential data processing. Their interest is due to a new way of HMM's posterior marginal distribution computation based on the Entropic Forward-Backward (EFB) algorithm, which allows considering arbitrary features \cite{azeraf2020hidden} with HMM. 
We adapt EFB to HMM of order 2 (HMM2) and HMM with complexified noise (HMM-CN), presented in the next section. Therefore, we present HNMC as the HMM neural extension, HNMC2 as the HMM2 one, and HNMC-CN as the HMM-CN one.

The paper is organized as follows. The next section presents the HMM model, its EFB algorithm, the HMM2, the HMM-CN, and their EFB algorithms. Then we introduce the HNMC, the HNMC2, and the HNMC-CN models. We specify the computational graph and related training process of the HNMC. We also describe the differences between our proposed models and some previous ones combining HMM and neural networks. The fourth part is devoted to experiments. We compare our models with RNN and Bidirectional RNN (BiRNN) \cite{schuster1997bidirectional} for different sequence labeling tasks: Part-Of-Speech (POS) tagging, Chunking, and Named-Entity-Recognition (NER). We implement many architectures with various embedding methods to reach a convincing empirical comparison. We only compare with RNN and BiRNN, as the latter's extensions to catch longer memory information, leading to LSTM and GRU, is discussed as the perspectives for HNMC based models in the last section.

\section{\uppercase{Hidden Markov Model}}

\subsection{Description of the HMM}

The Hidden Markov Model is a sequential model created sixty years ago and used in numerous applications \cite{rabiner1989tutorial,li2000image,brants-2000-tnt}. It allows the restoration of a hidden sequence from an observed one. 

Let $x_{1:T} = (x_1, ...,x_T)$ be a hidden realization of a stochastic process, taking its values in $\Lambda_X = \{ \lambda_1, ...,\lambda_N \}$, and let $y_{1:T}=(y_1, ..., y_T)$ be an observed realization of a stochastic one, taking its values in $\Omega_Y = \{ \omega_1, ...,\omega_M \}$.
The couple $(x_{1:T}, y_{1:T})$ is a HMM if its probabilistic law can be written:
\begin{align*}
p( x_{1:T}, y_{1:T} ) &= p(x_1) p(y_1 | x_1) p(x_2 | x_1) \\ &p(y_2 | x_2) ... p(x_T | x_{T-1}) p(y_T | x_T)
\end{align*}
The probabilistic oriented graph of the HMM is given in figure 1.

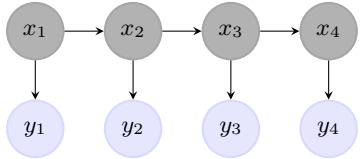
\begin{figure}
\begin{center}
\begin{tikzpicture}
[font=\small, inner sep=0pt, hidden/.style = {circle,draw = blue!15, fill = blue!10, thick,minimum size = 0.75cm, rounded corners}, visible/.style = {circle,draw=black!35,fill=black!30,thick,minimum size=0.75cm, rounded corners}, scale = 0.65]
\node at (-5,2) (x1) [visible] {$x_1$};
\node at (-5,0) (h1) [hidden]  {$y_1$};
\draw [->, >=stealth] (x1) to (h1);
                    
\node at (-3,2) (x2) [visible] {$x_2$};
\node at (-3,0) (h2) [hidden]  {$y_2$};
\draw [->, >=stealth] (x2) to (h2);
                    
\node at (-1,2) (x3) [visible] {$x_3$};
\node at (-1,0) (h3) [hidden]  {$y_3$};
\draw [->, >=stealth] (x3) to (h3);
                    
\node at (1,2) (x4) [visible] {$x_4$};
\node at (1,0) (h4) [hidden]  {$y_4$};
\draw [->, >=stealth] (x4) to (h4);
                    
\draw [->, >=stealth] (x1) to (x2);
\draw [->, >=stealth] (x2) to (x3);
\draw [->, >=stealth] (x3) to (x4);
\end{tikzpicture}
\captionof{figure}{Probabilistic oriented graph of the HMM}
\end{center}
\end{figure}

\subsection{The Entropic Forward-Backward algorithm for HMM}

There are different ways to restore a hidden chain from an observed one using the HMM. With the Maximum A Posteriori criterion (MAP), one can use the classic Viterbi \cite{viterbi1967error} algorithm. About the Maximum Posterior Mode (MPM), one can use the classic Forward-Backward \cite{rabiner1989tutorial} (FB) one. However, both Viterbi and FB algorithms use probabilities $p(y_t |x_t)$, making them impossible to consider arbitrary features of the observations \cite{jurafsky2000speech,sutton2006introduction}, especially the output of a neural network function. To correct this default, the Entropic Forward Backward (EFB) algorithm specified below computes the MPM using $p(x_t |y_t)$ and can take into account any features \cite{azeraf2020hidden}. This makes possible the neural extension of the HMM we are going to present.

For stationary HMM we consider in the whole paper, the EFB deals with the following parameters:
\begin{itemize}
    \item $\pi(i) = p(x_t = \lambda_i)$;
    \item $a_i(j) = p(x_{t+1} = \lambda_j | x_t = \lambda_i)$;
    \item $L_y(i) = p(x_{t} = \lambda_i | y_t = y)$;
\end{itemize}

The MPM restoration method we consider consists of maximization of the probabilities $p(x_t = \lambda_i | y_{1:T})$. They are given from entropic forward $\alpha$ and entropic backward $\beta$ functions with:
\begin{align}
p(x_t = \lambda_i | y_{1:T}) = \frac{\alpha_t(i) \beta_t(i)}{\sum_{j = 1}^N \alpha_t(j) \beta_t(j)}
\end{align}

Entropic forward functions are computed recursively as follows:
\begin{itemize}
\item For $t = 1$: 
\begin{align*}
\alpha_1(i) = L_{y_1}(i)
\end{align*}
\item For $1 \leq t < T$:
\begin{align}
\alpha_{t+1}(i) &= \frac{L_{y_{t+1}}(i)}{\pi(i)} \sum_{j = 1}^N \alpha_t(j) a_j(i)
\label{hmm_efb_alpha_rec}
\end{align}
\end{itemize}
And the entropic backward ones:
\begin{itemize}
\item For $t = T:$
\begin{align*}
\beta_T(i) = 1
\end{align*}
\item For $1 \leq t < T$:
\begin{align}
\beta_{t}(i) & = \sum_{j = 1}^N \frac{L_{y_{t+1}}(j)}{\pi(j)} \beta_{t+1}(j) a_{i}(j)
\label{hmm_efb_beta_rec}
\end{align}
\end{itemize}
One can normalize values at each time in (\ref{hmm_efb_alpha_rec}) and (\ref{hmm_efb_beta_rec}) to avoid underflow problems without modifying the probabilities' computation. 

\begin{figure}[t]
\begin{center}
\begin{tikzpicture}
[font=\small, inner sep=0pt, hidden/.style = {circle,draw = blue!15, fill = blue!10, thick,minimum size = 0.75cm, rounded corners}, visible/.style = {circle,draw=black!35,fill=black!30,thick,minimum size=0.75cm, rounded corners}, scale = 0.65]
\node at (-5,2) (x1) [visible] {$x_1$};
\node at (-5,0) (h1) [hidden]  {$y_1$};
\draw [->, >=stealth] (x1) to (h1);
                    
\node at (-3,2) (x2) [visible] {$x_2$};
\node at (-3,0) (h2) [hidden]  {$y_2$};
\draw [->, >=stealth] (x2) to (h2);
                    
\node at (-1,2) (x3) [visible] {$x_3$};
\node at (-1,0) (h3) [hidden]  {$y_3$};
\draw [->, >=stealth] (x3) to (h3);
                    
\node at (1,2) (x4) [visible] {$x_4$};
\node at (1,0) (h4) [hidden]  {$y_4$};
\draw [->, >=stealth] (x4) to (h4);
                    
\draw [->, >=stealth] (x1) to (x2);
\draw [->, >=stealth] (x2) to (x3);
\draw [->, >=stealth] (x3) to (x4);

\draw [->, >=stealth] (x1) to [out=45, in=135] (x3);
\draw [->, >=stealth] (x2) to [out=45, in=135] (x4);
\end{tikzpicture}
\captionof{figure}{Probabilistic oriented graph of the HMM of order 2}
\label{fig_hmm_2}
\end{center}
\end{figure}
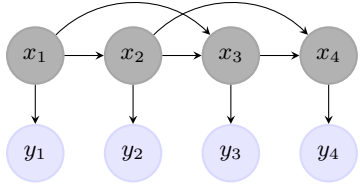

\subsection{EFB algorithm for HMM of order 2}

In this paragraph, we describe an extension of EFB above to HMM2, which allows to catch longer memory information than the HMM. The probabilistic law of $(x_{1:T}, y_{1:T})$ for the HMM2 is:
\begin{align*}
p(x_{1:T}, &y_{1:T}) = p(x_1) p(x_2 | x_1) p(x_3 | x_1, x_2) ... \\
& p(x_T | x_{T - 2}, x_{T - 1}) p(y_1 | x_1) p(y_2 | x_2) ... p(y_T | x_T)
\end{align*}
Its probabilistic graph is given in figure \ref{fig_hmm_2}.

We introduce the following notation to present the EFB algorithm for HMM2:
\begin{align*}
a^2_{i, j}(k) = p(x_{t + 2} = \lambda_k | x_t = \lambda_i, x_{t + 1} = \lambda_j )
\end{align*}

The EFB algorithm for HMM2 is the following:
\begin{itemize}
\item For $t = 1$:
\begin{align*}
p(x_1 = \lambda_i | y_{1:T}) = \frac{\sum_j \alpha_2^2(i, j) \beta_2^2(i, j)}{\sum_k \sum_j \alpha_2^2(k, j) \beta_2^2(k, j)}
\end{align*}

\item For $2 \leq t \leq T$:
\begin{align*}
p(x_t = \lambda_i | y_{1:T}) = \frac{\sum_j \alpha_t^2(j, i) \beta_t^2(j, i)}{\sum_k \sum_j \alpha_t^2(j, k) \beta_t^2(j, k)}
\end{align*}
\end{itemize}

The entropic forward-2 functions $\alpha^2$ are computed with the following recursion:
\begin{itemize}
\item For $t = 2$:
\begin{align*}
\alpha_2^2(j, i) = L_{y_1}(j) a_j(i) \frac{L_{y_2}(i)}{\pi(i)}
\end{align*}

\item And for $2 \leq t < T$:
\begin{align*}
\alpha_{t + 1}^2(j, i) = \sum_k \alpha_{t}^2(k, j) a^2_{k, j}(i) \frac{L_{y_{t + 1}}(i)}{\pi(i)}
\end{align*}
\end{itemize}

\noindent And the backward-2 functions $\beta^2$ with the following one:
\begin{itemize}
\item For $t = T$:
\begin{align*}
\beta_T^2(j, i) = 1
\end{align*}

\item And for $2 \leq t < T$:
\begin{align*}
\beta_{t}^2(j, i) = \sum_k \beta_{t + 1}^2(i, k) a^2_{j, i}(k) \frac{L_{y_{t + 1}}(k)}{\pi(k)}
\end{align*}
\end{itemize}

\subsection{HMM-CN and related EFB}

\begin{figure}[t]
\begin{center}
\begin{tikzpicture}
[font=\small, inner sep=0pt, hidden/.style = {circle,draw = blue!15, fill = blue!10, thick,minimum size = 0.75cm, rounded corners}, visible/.style = {circle,draw=black!35,fill=black!30,thick,minimum size=0.75cm, rounded corners}, scale = 0.65]
\node at (-5,2) (x1) [visible] {$x_1$};
\node at (-5,0) (h1) [hidden]  {$y_1$};
\draw [->, >=stealth] (x1) to (h1);
                    
\node at (-3,2) (x2) [visible] {$x_2$};
\node at (-3,0) (h2) [hidden]  {$y_2$};
\draw [->, >=stealth] (x2) to (h2);
                    
\node at (-1,2) (x3) [visible] {$x_3$};
\node at (-1,0) (h3) [hidden]  {$y_3$};
\draw [->, >=stealth] (x3) to (h3);
                    
\node at (1,2) (x4) [visible] {$x_4$};
\node at (1,0) (h4) [hidden]  {$y_4$};
\draw [->, >=stealth] (x4) to (h4);
                    
\draw [->, >=stealth] (x1) to (x2);
\draw [->, >=stealth] (x2) to (x3);
\draw [->, >=stealth] (x3) to (x4);

\draw [->, >=stealth] (x1) to (h2);
\draw [->, >=stealth] (x2) to (h1);
\draw [->, >=stealth] (x2) to (h3);
\draw [->, >=stealth] (x3) to (h2);
\draw [->, >=stealth] (x3) to (h4);
\draw [->, >=stealth] (x4) to (h3);
\end{tikzpicture}
\captionof{figure}{Probabilistic oriented graph of the HMM-CN}
\label{fig_hmm_in}
\end{center}
\end{figure}
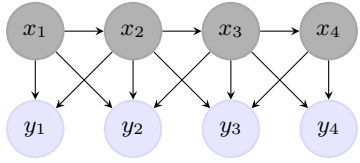

This paragraph describes the new HMM-CN model with related new EFB. It is another extension of HMM aiming to improve its results. Its probabilistic oriented graph is presented in figure \ref{fig_hmm_in}. 

In this case, the hidden sequence is still a Markov chain, and the conditional law of the observation $y_t$ given $x_{1:T}$ depends on $x_{t-1}, x_t$, and $x_{t + 1}$, implying stronger dependency with the hidden chain. The HMM-CN has the probabilistic law:
\begin{align*}
p(x_{1:T}, &y_{1:T}) = p(x_1) p(x_2 | x_1) p(x_3 | x_2) ... p(x_T | x_{T - 1}) \\
& p(y_1 | x_1, x_2) p(y_2 | x_1, x_2, x_3) ... p(y_T | y_{T - 1}, y_T)
\end{align*}

To present the EFB algorithm for HMM-CN, we set:
\begin{itemize}
\item $I_{j, y}(i) = p(x_{t + 1} = \lambda_i | x_t = \lambda_j, y_t = y)$
\item $J_{j, y}(i) = p(x_{t} = \lambda_i | x_{t+1} = \lambda_j, y_{t + 1} = y)$
\end{itemize}

The goal of the EFB algorithm is to compute $p(x_t = \lambda_i | y_{1:T})$, using $I_{j, y}(i)$ and $J_{j, y}(i)$ above, we show:
\begin{align*}
p(x_t = \lambda_i | y_{1:T}) = \frac{\alpha^{CN}_t(i) \beta^{CN}_t(i)}{\sum_j \alpha^{CN}_t(j) \beta^{CN}_t(j)},
\end{align*}
with the entropic forward-cn functions $\alpha^{CN}$ computed with the following recursion:
\begin{itemize}
\item For $t = 1$:
\begin{align*}
\alpha^{CN}_1(i) = L_{y_1}(i)
\end{align*}
\item And for $1 \leq t < T$:
\begin{align}
\alpha^{CN}_{t + 1}(i) = \sum_j \alpha^{CN}_t(j) I_{j, y_t}(i) \frac{L_{y_{t+1}}(i) J_{i, y_{t+1}}(j)}{\pi(j) a_j(i)} \label{alpha_hmmin_rec}
\end{align}
\end{itemize}

\noindent
And the entropic backward-cn functions $\beta^{CN}$ computed with the following one:
\begin{itemize}
\item For $t = T$:
\begin{align*}
\beta^{CN}_T(i) = 1
\end{align*}
\item And for $1 \leq t < T$:
\begin{align}
\beta^{CN}_{t}(i) = \sum_j \beta^{CN}_{t + 1}(j) I_{i, y_t}(j) \frac{L_{y_{t+1}}(j) J_{j, y_{t+1}}(i)}{\pi(i) a_i(j)} \label{beta_hmmin_rec}
\end{align}
\end{itemize}

Proofs of the EFB algorithms for HMM2 and HMM-CN are in the appendixes.

\section{\uppercase{Hidden Neural Markov Chain Framework}}

\subsection{Construction of the HNMC}

To extend the HMM considered above to the HNMC, we have to model the three functions, $\pi$, $a$, and $L$, with a feedforward neural network function modeling $\frac{L_{y_{t + 1}}(i)}{\pi(i)} a_j(i)$. This neural network function has $y_{t + 1}$ concatenated with the one-hot encoding of $j$ as input, and outputs a positive vector of size $N$. To do that, we use a last positive activation function as the exponential, the sigmoid, or a modified Exponential Linear Unit (mELU):
\begin{align*}
f(x) = \left\{
    \begin{array}{ll}
        1 + x & \mbox{if } x > 0 \\
        e^x & \mbox{otherwise.}
    \end{array}
\right.
\end{align*}
Then, we apply the EFB algorithm for sequence restoration. The first step of the algorithm is performed thanks to the introduction of an initial state, which can be drawn randomly or equals to a constant different from $0$. Therefore, we have constructed the HNMC, a new model able to process sequential data in a “term-to-term” way with neural network functions.

\begin{figure}
\begin{center}
\begin{tikzpicture}
[font=\small, inner sep=0pt, hidden/.style = {circle,draw = blue!15, fill = blue!10, thick,minimum size = 0.75cm, rounded corners}, interior/.style = {circle,draw = yellow!50, fill = yellow!50, thick,minimum size = 0.65cm, rounded corners}, visible/.style = {circle,draw=black!35,fill=black!30,thick,minimum size=0.75cm, rounded corners}, scale = 0.65, square/.style={regular polygon,regular polygon sides=4}]

\node at (1,-1) (y1) [hidden] {$y_1$};
\node at (3,-1) (y2) [hidden] {$y_2$};
\node at (5,-1) (y3) [hidden] {$y_3$};
\node at (7,-1) (y4) [hidden] {$y_4$};
\node at (9,-1) (y5) [hidden] {$y_5$};

\draw[rounded corners=5pt, draw = black!50, fill = black!15] (0,0) rectangle ++(10,1.25);

\node at (1,0.625) (h11) [interior] {$h_1^{(1)}$};
\node at (3,0.625) (h21) [interior] {$h_2^{(1)}$};
\node at (5,0.625) (h31) [interior] {$h_3^{(1)}$};
\node at (7,0.625) (h41) [interior] {$h_4^{(1)}$};
\node at (9,0.625) (h51) [interior] {$h_5^{(1)}$};

\draw[rounded corners=5pt, draw = black!50, fill = black!15] (0,1.75) rectangle ++(10,1.25);

\node at (1,2.375) (h12) [interior] {$h_1^{(2)}$};
\node at (3,2.375) (h22) [interior] {$h_2^{(2)}$};
\node at (5,2.375) (h32) [interior] {$h_3^{(2)}$};
\node at (7,2.375) (h42) [interior] {$h_4^{(2)}$};
\node at (9,2.375) (h52) [interior] {$h_5^{(2)}$};



\node at (1,4.125) (x1) [visible] {$x_1$};
\node at (3,4.125) (x2) [visible] {$x_2$};
\node at (5,4.125) (x3) [visible] {$x_3$};
\node at (7,4.125) (x4) [visible] {$x_4$};
\node at (9,4.125) (x5) [visible] {$x_5$};

\draw [<->, >=stealth] (h11) to (h21);
\draw [<->, >=stealth] (h21) to (h31);
\draw [<->, >=stealth] (h31) to (h41);
\draw [<->, >=stealth] (h41) to (h51);

\draw [<->, >=stealth] (h12) to (h22);
\draw [<->, >=stealth] (h22) to (h32);
\draw [<->, >=stealth] (h32) to (h42);
\draw [<->, >=stealth] (h42) to (h52);


\draw [->, >=stealth] (y1) to (h11);
\draw [->, >=stealth] (y2) to (h21);
\draw [->, >=stealth] (y3) to (h31);
\draw [->, >=stealth] (y4) to (h41);
\draw [->, >=stealth] (y5) to (h51);

\draw [->, >=stealth] (h11) to (h12);
\draw [->, >=stealth] (h21) to (h22);
\draw [->, >=stealth] (h31) to (h32);
\draw [->, >=stealth] (h41) to (h42);
\draw [->, >=stealth] (h51) to (h52);


\draw [<-, >=stealth] (x1) to (h12);
\draw [<-, >=stealth] (x2) to (h22);
\draw [<-, >=stealth] (x3) to (h32);
\draw [<-, >=stealth] (x4) to (h42);
\draw [<-, >=stealth] (x5) to (h52);
\end{tikzpicture}
\captionof{figure}{Computational graph of the HNMC with two hidden layers}
\label{fig_hnmc_2_layers}
\end{center}
\end{figure}
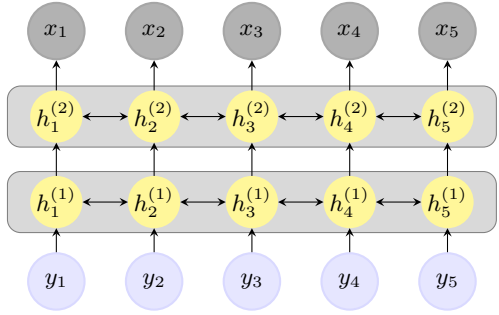

We can stack HNMCs to add hidden layers, similarly to the stacked RNN practice, to achieve greater model complexity. The output of a first HNMC based EFB restoration layer becoming the input of the next one, and so on, applying the EFB layer after layer.  For example, a computational graph of a HNMC composed of four layers is specified in figure \ref{fig_hnmc_2_layers}. In the general case, we have $K + 2$ layers: 
\begin{itemize}
    \item An input layer $y$;
    \item $K$ hidden layers $h^{(1)}, h^{(2)}, \ldots, h^{(K)}$;
    \item An output layer $x$.
\end{itemize}

We consider that ${(H}^{(1)},\ Y)$, ${(H}^{(2)},\ H^{(1)})$, ..., ${(H}^{(K)},\ H^{(K-1)})$, are HMMs, and the last layer $H^{(K)}$ is connected with the output layer $x$ thanks to a feedfoward neural network function denoted $f$. Finally, we compute for each $t \in \left\{ 1, \ldots, T \right\}$, $x_t$ from $y_{1:T}$ as follows:
\begin{enumerate}
    \item Computing $h^{(1)}$ from $y_{1:T}$ using EFB;
	\item Computing $h^{(2)}$ from $h^{(1)}$ using EFB, considering $h^{(1)}$ as the observations; then compute $h^{(3)}$ from $h^{(2)}$ using EFB, ...
	\item Computing $h^{(K)}$ from $h^{(K-1)}$ using EFB, considering $h^{(K-1)}$ as the observations;
	\item Computing $x_t = f(h_t^{(K)})$, $x_t$ is the output vector of probabilities of the different states at time $t$.
\end{enumerate}

Let us notice that, from a probabilistic point of view, this stacked HNMC can be seen as a particular Triplet Markov Chain \cite{pieczynski2003triplet} having $K + 2$ layers, and our restoration method would be an approximation of this model.

Thus, the HNMC can be used as a sequential neural model with term-to-term processing, like the RNN. However, unlike the latter, the HNMC uses all the observation $y_{1:T}$ to restore $x_t$, whereas the RNN uses only $y_{1:t}$. One can use the BiRNN to correct this default, consisting of applying a RNN from right to left, another one from left to right, then concatenate the outputs.

\subsection{Neural extension of HMM2 and HMM-CN}

Neural extensions of HMM2 and HMM-CN follow the same principles as for HMM.
For the HMM2, we model $a^2_{k, j}(i) \frac{L_{y_{t + 1}}(i)}{\pi(i)}$ with a feedfoward neural functions with a positive last activation function, taking as input $y_{t + 1}$ and the one-hot encoding of $(k, j)$. This model is denoted HMNC2.

Concerning the HMM-CN, we use two different neural functions: one to model $\frac{J_{i, y_{t+1}}(j)}{\pi(j)}$, and the other one to model $I_{j, y_t}(i) \frac{L_{y_{t+1}}(i)}{a_j(i)}$, with the relevant inputs, and positive outputs. This model is denoted HMNC-CN.

\subsection{Learning HNMC based models' parameters}

To learn the different parameters of each of our new models, we consider the backpropagation algorithm \cite{lecun1988theoretical,lecun1989backpropagation} frequently used for neural network learning. Given a loss, for example the cross-entropy $L_{CE}$, $\theta$ a parameter of one of the model’s functions, and a sequence $y_{1:T}$, we compute $\frac{\partial L_{CE}}{\partial \theta}$ with gradient backpropagation over all the intermediary variables.
Then, we apply the gradient descent \cite{ruder2016overview} algorithm:
\begin{align*}
\theta^{(new)} = \theta - \kappa \frac{\partial L_{CE}}{\partial \theta}
\end{align*}
with $\kappa$ the learning rate.

As for any neural network architectures, we can apply the gradient descent algorithm for HNMC based models. Therefore, we can create different architectures and combine them with other neural network models as Convolutional Neural Networks \cite{lecun1999object} or feedforward ones.

\subsection{Related works}

The combination of HMM with neural networks starts in the 1990s \cite{bengio1994globally}, focusing on the concatenation of the two models. Nowa
few papers deal with the subject. The closest model to HNMC is the neural HMM proposed in \cite{tran-etal-2016-unsupervised}. However, the proposed method is not EFB based, and neural networks model different parameters from those considered in this paper. Indeed, they model $p(y_t|{x_t=\lambda}_i)$. This implies a sum over all the possible observations to be computed, which considerably increases the number of parameters for NLP applications, where observations are words. It also avoids the combination with embedding methods, aiming to convert a word into a continuous vector. Moreover, the proposed training method is based on the Baum-Welch algorithm with Expectation-Maximization \cite{welch2003hidden}, or Direct Marginal Likelihood \cite{salakhutdinov2003optimization}, so the ability to create various architectures as it is done with RNN is not trivial. It focuses on unsupervised tasks, which is not the case for HNMC. Comparable works can be found in \cite{wang2017hybrid,wang2018neural}. Therefore, the proposed HNMC, based on different neuralized parameters with gradient descent training and aiming a different objective, is an original way to combine HMM with neural networks.

\section{\uppercase{Experiments}}

\begin{table*}[t]
\centering
\begin{tabular}{l|cc|ccc}
\hline
\multicolumn{6}{c}{\textbf{Architecture 1}} \\
\hline
\hline
{} & RNN & BiRNN & HNMC & HNMC2 & HNMC-CN \\
\hline
POS Ext UD & $88.40\% \pm 0.02$ & $91.38\% \pm 0.04$ & $90.98\% \pm 0.03$ & $91.33\% \pm 0.04$ & $\bm{92.62\%} \pm 0.04$ \\
Ch GloVe 00 & $86.68 \pm 0.08$ & $90.76 \pm 0.55$ & $87.77 \pm 0.13$ & $88.18 \pm 0.04$ & $\bm{92.02} \pm 0.03$ \\
NER FT 03 & $81.91 \pm 0.14$ & $82.62 \pm 0.56$ & $83.41 \pm 0.10$ & $83.49 \pm 0.06$ & $\bm{87.49} \pm 0.19$ \\
\hline
\end{tabular}
\caption{\label{results-table-1} Results of the different models for POS Tagging, Chunking, and NER, for the Architecture 1 - the model only.
}
\end{table*}

\begin{table*}[t]
\centering
\begin{tabular}{l|cc|ccc|c}
\hline
\multicolumn{7}{c}{\textbf{Architecture 2}} \\
\hline
\hline
{} & RNN & BiRNN & HNMC & HNMC2 & HNMC-CN & HS \\
\hline
POS Ext UD & $89.84\% \pm 0.04$ & $93.07\% \pm 0.05$ & $92.77\% \pm 0.06$ & $93.01\% \pm 0.04$ & $\bm{93.29\%} \pm 0.05$ & $50$ \\
Ch GloVe 00 & $93.85 \pm 0.06$ & $95.02 \pm 0.11$ & $95.43 \pm 0.09$ & $\bm{95.59} \pm 0.13$ & $95.36 \pm 0.07$ & $32$ \\
NER FT 03 & $84.53 \pm 0.21$ & $87.52 \pm 0.13$ & $88.22 \pm 0.13$ & $88.47 \pm 0.05$ & $\bm{89.40} \pm 0.03$ & $20$ \\
\hline
\end{tabular}
\caption{\label{results-table-2} Results of the different models for POS Tagging, Chunking, and NER, for the Architecture 2 - the model followed by a feedforward neural function, the hidden size is denoted HS.
}
\end{table*}

\begin{table*}[t!]
\centering
\begin{tabular}{l|cc|ccc|c}
\hline
\multicolumn{7}{c}{\textbf{Architecture 3}} \\
\hline
\hline
{} & RNN & BiRNN & HNMC & HNMC2 & HNMC-CN & HS \\
\hline
POS Ext UD & $89.20\% \pm 0.09$ & $92.80\% \pm 0.21$ & $92.73\% \pm 0.12$ & $92.97\% \pm 0.08$ & $\bm{93.36\%} \pm 0.03$ & $50$ \\
Ch GloVe 00 & $93.13 \pm 0.14$ & $94.91 \pm 0.09$ & $95.53 \pm 0.13$ & $\bm{95.59} \pm 0.06$ & $95.40 \pm 0.14$ & $32$ \\
NER FT 03 & $85.10 \pm 0.12$ & $88.68 \pm 0.31$ & $88.02 \pm 0.19$ & $88.66 \pm 0.33$ & $\bm{89.37} \pm 0.12$ & $20$ \\
\hline
\end{tabular}
\caption{\label{results-table-3} Results of the different models for POS Tagging, Chunking, and NER, for the Architecture 3 - two models stacked, the hidden size is denoted HS.
}
\end{table*}

This section presents some experimental results comparing the RNN, the BiRNN, the HNMC, the HNMC2, and the HNMC-CN. After some preliminary presentations of the different tasks and the word embedding process, we create different architectures for all the models and test them for sequence labeling applications. Motivations to the choice of comparing our models with RNN and BiRNN are discussed in perspectives. 

\subsection{Sequence labeling tasks}

We select sequence labeling applications as they are the most intuitive tasks to apply a sequential model in the NLP framework. It consists of labeling every word in a sentence with a specific tag. We apply the different models to POS Tagging, Chunking, and NER, which are among the most popular sequence labeling applications.

The POS tagging consists of labeling every word with its grammatical function as noun (\textit{NOUN}), verb (\textit{VERB}), determinant (\textit{DET}), etc. For example, the sentence \textit{(Batman, is, the, vigilante, of, Gotham, .)} has the labels \textit{(NOUN, VERB, DET, NOUN, PREP, NOUN, PUNCT)}. The accuracy score is used to evaluate this task.

Chunking consists of segmenting a sentence with a more global point of view than the POS tagging. It decomposes the sentence by groups of words linked by a syntactic function, as a noun phrase (\textit{NP}), a verb phrase (\textit{VP}), an adjective phrase (\textit{ADJP}), among others. For example, the sentence \textit{(The, worst, enemy, of, Batman, is, the, Joker, .)} has the following chunk tags \textit{(NP, NP, NP, PP, NP, VP, NP, NP, O)}. \textit{O} denotes a word having no chunk tag. The $F_1$ score is used to measure the performance of this task.

The objective of the NER is to find the different entities in a sentence. Entities can be the name of a person (\textit{PER}), of a city (\textit{LOC}), or of a company (\textit{ORG}). For example, the sentence \textit{(Bruce, Wayne, ,, a, citizen, of, Gotham, ,, is, the, secret, identity, of, Batman, .)} can have the entities \textit{(PER, PER, O, O, O, O, LOC, O, O, O, O, O, O, PER, O)}. The entity set depends on the use-case, and one can it change according to the objective.
As for Chunking, the $F_1$ score is used to evaluate the performances of a model. 

For our experiments, we use three reference datasets: Universal Dependencies English (UD En) \cite{nivre2016universal} for POS Tagging, CoNLL 2000 \cite{tjong-kim-sang-buchholz-2000-introduction} for Chunking, and we use general entites with the CoNLL 2003 \cite{tjong-kim-sang-de-meulder-2003-introduction} dataset for NER\footnote{All these datasets are freely available: UD En on the website \href{https:/universaldependencies.org/\#language-}{https:/universaldependencies.org/\#language-}, CoNLL 2000, for example, with NLTK \cite{loper2002nltk} library, and CoNLL 2003 after a demand on \href{https:/www.clips.uantwerpen.be/conll2003//ner/}{https:/www.clips.uantwerpen.be/conll2003//ner/}}.

\subsection{Word Embedding methods}

A sentence is composed of textual data; this type of data cannot be the input of feedforward neural network functions. Indeed, these functions have as input a numerical vector or scalar. Our experiments' first step consists of a pre-processing task to convert a word into a numerical vector, called word embedding, or word encoding.
In order to make our conclusions independent from embedding, we use three different embedding methods: GloVe \cite{pennington2014glove}, FastText \cite{bojanowski2017enriching}, and EXT encoding \cite{komninos-manandhar-2016-dependency}.

\subsection{The different architectures}

To compare the different models for the different sequence labeling tasks, we implement three architectures for each model:
\begin{itemize}
\item Architecture 1: only the model;
\item Architecture 2: the model followed with a feedforward neural network function, equivalent of the figure \ref{fig_hnmc_2_layers} with the layers $(y, h^{(1)}, x)$ for the HNMC;
\item Architecture 3: two models stacked, equivalent of the figure \ref{fig_hnmc_2_layers} with the layers $(y, h^{(1)}, h^{(2)})$ for the HNMC.
\end{itemize}

\subsection{Experimental details}

Every model is programmed in python using PyTorch \cite{paszke2019pytorch} library for automatic differentiation, and Flair library \cite{akbik2019flair} for word encoding. 
The loss function is the cross-entropy. All the different parameters are modeled with feedforward neural networks without hidden layers, equivalent to the logistic regression.
About the activation functions, the HNMC based models always use mELU. For the RNN and BiRNN, we use them as usual, with hyperbolic tangent functions. Every model uses the softmax function at the end of the architecture to output probabilities.
We use Adam optimizer \cite{kingma2014adam} for all experiments, with a mini-batch size of $32$. For architecture 1, the learning rate equals $0.005$. We use different learning rates for the different layers for the other architectures: $0.05$, then $0.005$. This configuration gives the best experimental results for every model.

\subsection{Results}

For each architecture, we realize three experiments: POS Tagging with UD En using EXT (POS Ext UD), Chunking with CoNLL 2000 using GloVe (Ch GloVe 00), and NER with CoNLL 2003 using FastText (NER FT 03). Each experiment is done five times; we report the mean and the 95\%-confidence interval in Table \ref{results-table-1}, Table \ref{results-table-2}, and Table \ref{results-table-3}, with the different sizes of hidden layers, denoted HS.

First of all, we can notice that HNMC is always better than RNN. It is certainly because HNMC uses all the observations to restore any hidden variable, making it a bidirectional alternative to the RNN without increasing the number of parameters, which are slightly equivalent.
As expected, the HNMC2 achieves better results than the HNMC, and therefore the RNN. However, HNMC2 does not reach BiRNN scores, except in some cases, especially for Chunking.

Another interesting comparison concerns HNMC-CN and BiRNN. Indeed, the HNMC-CN achieves better results than the BiRNN for every experiment. 
It is a promising result, as prevalent models as BiLSTM and BiGRU are based on the BiRNN. Therefore, the HNMC-CN can be an alternative to the BiRNN for sequence labeling applications. These different results, comparing HNMC based models with RNN and BiRNN, show the proposed sequential neural framework’s potential.

\section{Conclusion and perspectives}

We have presented the HNMC framework, a new family a sequential neural models, introducing the classic HNMC, the HNMC2, and the HNMC-CN. We have compared these three models with the RNN and the BiRNN ones. On the one hand, the HNMC achieves better results than the RNN with an equivalent number of parameters. On the other hand,  the HNMC-CN has achieved better results than BiRNN for the different sequence labeling tasks. 

As a promising perspective, we can extend the HNMC-CN with long-memory methods, as BiRNN is extended to BiLSTM and BiGRU. Therefore, these extensions of HNMC-CN are expected to compete with BiLSTM and BiGRU. It is a challenging perspective, as these models are the most prevalent ones for sequential data processing.

\bibliographystyle{unsrt}  
\bibliography{references} 

\section*{\uppercase{Appendix}}

We introduce a new notation, for each $t \in \{ 1, ..., T \}, \lambda_i \in \Lambda_X, y_t \in \Omega_Y$: $b_i(y_t) = p(y_t | x_t = \lambda_i)$.

\subsection*{Proof of the EFB algorithm for HMM2}

The EFB algorithm for HMM2 aims to compute, for each $t \in \{ 1, ..., T \}, \lambda_i \in \Lambda_X, p(x_t = \lambda_i | y_{1:T})$. \\
We can show, with the law of HMM2 and figure \ref{fig_hmm_2}, for each $t > 1$:
\begin{align*}
p(x_t = i | y_{1:T}) = \frac{\sum_j \alpha^{2'}_t(j, i) \beta^{2'}_t(j, i) }{\sum_k \sum_j \alpha^{2'}_t(j, k) \beta^{2'}_t(j, k) }
\end{align*}
with:
\begin{align*}
\alpha^{2'}_t(j, i) &= p(x_{t - 1} = \lambda_j, x_t = \lambda_i, y_{1:t}) \\
\beta^{2'}_t(j, i) &= p(y_{t + 1:T} | x_{t - 1} = \lambda_j, x_t = \lambda_i)
\end{align*}

$\alpha^{2'}$ can be computed with the following recursion:
\begin{itemize}
\item For $t = 2$: $\alpha_2^{2'}(j, i) = \pi(j) b_j(y_1) a_j(i) b_i(y_2)$
\item For $2 \leq t < T$:
\begin{align*}
\alpha_{t + 1}^{2'}(j, i) = \sum_k \alpha_t^{2'}(k, j) a^2_{k, j}(i) b_i(y_{t + 1})
\end{align*}
\end{itemize}

\noindent
And $\beta^{2'}$ with the following one:
\begin{itemize}
\item For $t = T, \beta_T^{2'}(j, i) = 1$ 
\item For $2 \leq t < T$:
\begin{align*}
\beta_{t}^{2'}(j, i) &= \sum_k \beta^{2'}_{t + 1}(i, k) a^2_{j, i}(k) b_k(y_{t + 1})
\end{align*}
\end{itemize}

We can show, for each $2 \leq t \leq T$:
\begin{align}
\alpha_t^{2}(j, i) &= \frac{\alpha_t^{2'}(j, i)}{p(y_1) p(y_2) ... p(y_t)} \label{eq_alpha_2} \\
\beta_t^{2}(j, i) &= \frac{\beta_t^{2'}(j, i)}{p(y_{t + 1}) p(y_{t + 2}) ... p(y_T)}
\label{eq_beta_2}
\end{align}

For $t = 2$,
\begin{align*}
\alpha_2^{2'}(j, i) &= p(y_1, x_1 = \lambda_j, y_2, x_2 = \lambda_i) \\
&= p(y_1) L_{y_1}(j) p(y_2) a_j(i) \frac{L_{y_2}(i)}{\pi(i)}
\end{align*}
Therefore, (\ref{eq_alpha_2}) is true for $t = 2$. We suppose (\ref{eq_alpha_2}) for $t$, and we prove it for $t + 1$:
\begin{align*}
\alpha_{t + 1}^{2}(j, i) &= \sum_k \frac{\alpha_t^{2'}(k, j)}{p(y_1) p(y_2) ... p(y_t)} a^2_{k, j}(i) \frac{b_i(y_{t + 1})}{p(y_{t + 1})} \\
&= \frac{\alpha_{t + 1}^{2'}(j, i)}{p(y_1) p(y_2) ... p(y_t) p(y_{t + 1})}
\end{align*}

\noindent
About (\ref{eq_beta_2}), it is true for $t = T$. We suppose (\ref{eq_beta_2}) true for $t + 1 < T$, and we prove it for $t$:
\begin{align*}
\beta_{t}^2(j, i) &= \sum_k \frac{\beta_{t + 1}^{2'}(i, k)}{p(y_{t + 2}) ... p(y_T)} a^2_{j, i}(k) \frac{b_k(y_{t + 1})}{p(y_{t + 1})} \\
&= \frac{\beta_{t}^{2'}(j, i)}{p(y_{t + 1}) p(y_{t + 2}) ... p(y_T)}
\end{align*}

\noindent
(\ref{eq_alpha_2}) and (\ref{eq_beta_2}) and proved for each $t$. \\
Therefore,
\begin{align*}
p(x_t = \lambda_i | y_{1:T}) = \frac{\alpha_t^{2'}(i) \beta_t^{2'}(i)}{\sum_j \alpha_t^{2'}(j) \beta_t^{2'}(j)} = \frac{\alpha_t^{2}(i) \beta_t^{2}(i)}{\sum_j \alpha_t^{2}(j) \beta_t^{2}(j)}
\end{align*}
Which ends the proof of EFB algorithm for HMM2.

\subsection*{Proof of the EFB algorithm for HMM-CN}

According to figure \ref{fig_hmm_in}, $(x_{1:t-1}, y_{1:t-1})$ and $(x_{t+1:T}, y_{t+1:T})$ are independent conditionally on $(x_t, y_t)$, and thus we have:

\begin{align*}
p(x_t = \lambda_i | y_{1:T}) & = \frac{\alpha_t^{CN'}(i) \beta_t^{CN'}(i)}{\sum_j \alpha_t^{CN'}(j) \beta_t^{CN'}(j)}
\end{align*}
with:
\begin{align*}
\alpha_t^{CN'}(i) &= p(x_t = \lambda_i, y_{1:t}) \\
\beta_t^{CN'}(i) &= p(y_{t+1 : T} | x_t = \lambda_i, y_t)
\end{align*}

\noindent
$\alpha^{CN'}$ can be computed with the following recursion:
\begin{itemize}
\item For $t = 1, \alpha_1^{CN'}(i) = \pi(i) p(y_t | x_1 = \lambda_i)$ 
\item For $1 \leq t < T$:
\begin{align*}
\alpha_{t + 1}^{CN'}(i) = \sum_j \alpha_{t}^{CN'}(j) I_{j, y_t}(i) p(y_{t + 1} | x_t = \lambda_j, x_{t + 1} = \lambda_i)
\end{align*}
\end{itemize}
And $\beta^{CN'}$ with the following one:
\begin{itemize}
\item For $t = T, \beta_T^{CN'}(i) = 1$ 
\item For $1 \leq t < T$:
\begin{align*}
\beta_{t}^{CN'}(i) &= \sum_j I_{i, y_t}(j) p(y_{t + 1} | x_t = \lambda_i, x_{t + 1} = \lambda_j) \beta_{t + 1}^{CN'}(j)
\end{align*}
\end{itemize}

We can show:
\begin{align}
\alpha_t^{CN}(i) &= \frac{\alpha_t^{CN'}(i)}{p(y_1) p(y_2) ... p(y_t)} \label{eq_alpha_in} \\
\beta_t^{CN}(i) &= \frac{\beta_t^{CN'}(i)}{p(y_{t + 1}) p(y_{t + 1}) ... p(y_T)}
\label{eq_beta_in}
\end{align}

\noindent
(\ref{eq_alpha_in}) is true for $t = 1$. We suppose (\ref{eq_alpha_in}) true for $t$, and we prove it for $t + 1$:
\begin{align*}
\alpha^{CN}_{t + 1}(i) &= \frac{1}{p(y_1) ... p(y_t)} \sum_j \alpha^{CN'}_t(j) I_{j, y_t}(i) \times \\
& \frac{p(x_{t + 1} = i | y_{t + 1}) p(x_{t} = \lambda_j | x_{t+1} = \lambda_i, y_{t + 1})}{p(x_t = \lambda_j, x_{t + 1} = \lambda_i)} \\
&= \frac{1}{p(y_1) ... p(y_t) p(y_{t + 1})} \sum_j \alpha^{CN'}_t(j) I_{j, y_t}(i) \times \\
& \frac{p(x_t = \lambda_j, x_{t + 1} = \lambda_i, y_{t + 1})}{p(x_t = \lambda_j, x_{t + 1} = \lambda_i)} \\
& = \frac{\alpha^{CN'}_{t + 1}(i)}{p(y_1) ... p(y_t) p(y_{t + 1})}
\end{align*}
Therefore, (\ref{eq_alpha_in}) is proved for all $t$. \\
The proof of (\ref{eq_beta_in}) follows the same reasoning. (\ref{eq_beta_in}) is true for $t = T$. We suppose (\ref{eq_beta_in}) true for $t + 1$, and we prove it a $t$:
\begin{align*}
\beta^{CN}_{t}(i) &= \frac{1}{p(y_{t + 2}) ... p(y_T)} \sum_j \beta^{CN'}_{t + 1}(j) I_{i, y_t}(j) \times \\
& \frac{p(x_{t + 1} = \lambda_j | y_{t + 1}) p(x_t = \lambda_i | x_{t + 1} = \lambda_j, y_{t + 1})}{p(x_t = \lambda_i, x_{t + 1} = \lambda_j)} \\
& = \frac{1}{p(y_{t + 1}) p(y_{t + 2}) ... p(y_T)} \sum_j \beta^{CN'}_{t + 1}(j) I_{i, y_t}(j) \times \\
& \frac{p(x_t = \lambda_i, x_{t + 1} = \lambda_j, y_{t + 1})}{p(x_t = \lambda_i, x_{t + 1} = \lambda_j)} \\
& = \frac{\beta^{CN'}_{t}(i)}{p(y_{t + 1}) p(y_{t + 2}) ... p(y_T)}
\end{align*}
, which prove (\ref{eq_beta_in}) for all $t$. \\
Therefore,
\begin{align*}
p(x_t = \lambda_i | y_{1:T}) = \frac{\alpha_t^{CN}(i) \beta_t^{CN}(i)}{\sum_j \alpha_t^{CN}(j) \beta_t^{CN}(j)}
\end{align*}
Which ends the proof of EFB algorithm for HMM-CN.

\end{document}